\documentclass[sigconf]{acmart}
\usepackage{cases}
\usepackage{amsmath} 
\usepackage{times}  
\usepackage{helvet}  
\usepackage{placeins}
\usepackage{courier}  
\usepackage{makecell}
\usepackage{subfigure}
\usepackage{algorithm}
\usepackage{algorithmic}
\usepackage[switch]{lineno}
\usepackage{multirow}
\usepackage{graphicx}
\usepackage{newfloat}
\usepackage{listings}
\AtBeginDocument{%
  \providecommand\BibTeX{{%
    \normalfont B\kern-0.5em{\scshape i\kern-0.25em b}\kern-0.8em\TeX}}}

\setcopyright{acmcopyright}

\copyrightyear{2023}
\acmYear{2023}
\setcopyright{acmlicensed}\acmConference[WWW '23 Companion]{Companion Proceedings of the ACM Web Conference 2023}{April 30-May 4, 2023}{Austin, TX, USA}
\acmBooktitle{Companion Proceedings of the ACM Web Conference 2023 (WWW '23 Companion), April 30-May 4, 2023, Austin, TX, USA}
\acmPrice{15.00}
\acmDOI{10.1145/3543873.3587299}
\acmISBN{978-1-4503-9419-2/23/04}

\begin{document}
\begin{sloppypar}

 \title{Graph-Level Embedding for Time-Evolving Graphs}

\author{Lili Wang}
\affiliation{%
  \institution{Dartmouth College}
  \city{Hanover}
  \state{New Hampshire}
  \country{USA}}
\email{lili.wang.gr@dartmouth.edu}

\author{Chenghan Huang}
\affiliation{%
  \institution{Millennium Management, LLC}
  \city{New York}
  \state{New York}
  \country{USA}}
\email{njhuangchenghan@gmail.com}

\author{Weicheng Ma}
\affiliation{%
  \institution{Dartmouth College}
  \city{Hanover}
  \state{New Hampshire}
  \country{USA}}
\email{weicheng.ma.gr@dartmouth.edu}

\author{Xinyuan Cao}
\affiliation{%
  \institution{Georgia Institute of Technology	}
  \city{Atlanta}
  \state{Georgia}
  \country{USA}}
\email{xcao78@gatech.edu}

\author{Soroush Vosoughi}
\affiliation{%
  \institution{Dartmouth College}
  \city{Hanover}
  \state{New Hampshire}
  \country{USA}}
\email{soroush.vosoughi@dartmouth.edu}

\renewcommand{\shortauthors}{Anonymous authors et al.}

\begin{abstract}

Graph representation learning (also known as network embedding) has been extensively researched with varying levels of granularity, ranging from nodes to graphs. While most prior work in this area focuses on node-level representation, limited research has been conducted on graph-level embedding, particularly for dynamic or temporal networks. However, learning low-dimensional graph-level representations for dynamic networks is critical for various downstream graph retrieval tasks such as temporal graph similarity ranking, temporal graph isomorphism, and anomaly detection. In this paper, we present a novel method for temporal graph-level embedding that addresses this gap. Our approach involves constructing a multilayer graph and using a modified random walk with temporal backtracking to generate temporal contexts for the graph's nodes. We then train a ``document-level’’ language model on these contexts to generate graph-level embeddings. We evaluate our proposed model on five publicly available datasets for the task of temporal graph similarity ranking, and our model outperforms baseline methods. Our experimental results demonstrate the effectiveness of our method in generating graph-level embeddings for dynamic networks.

\end{abstract}

\ccsdesc[300]{Computing methodologies~Learning latent representations}
\ccsdesc[500]{Networks~Topology analysis and generation}

\keywords{Dynamic Graph Embedding, Graph Representation Learning, Temporal Graphs, Graph Retrieval}

\maketitle

\section{Introduction}
Graphs, or networks, are prevalent in diverse domains such as social networks, protein interactions, and scientific collaboration. Graph representation learning, also known as graph embedding, enables the representation of graphs using general-purpose vector representations, removing the need for task-specific feature engineering. 

Graphs can be static, where their structure does not change over time, or dynamic, where their structure evolves over time. Social networks are typically dynamic due to their constantly changing structure. Representation learning on static and dynamic networks differs as static embeddings only need to capture network structure while dynamic embeddings must capture both structural and temporal aspects. While static embedding methods can be applied to dynamic networks, the resulting embeddings do not capture the evolving aspect of these networks. Network embedding methods are categorized by granularity, from node to graph level. Node embedding is the most common method in which nodes in a given network are represented as fixed-length vectors. While these vectors preserve different scales of proximity between the nodes, such as microscopic \cite{deepwalk,node2vec,wang2021embedding} and structural role \cite{struc2vec,graphwave,wang2020embedding,wang2021stress}, they cannot capture proximity between different networks as node representations are learned within the context of the network they occupy. Notably, considerable work has been done on node embedding for dynamic graphs \cite{dyngem,DDNE,wang2021hyperbolic,wang_tois_2021}, which preserves not only the network structural information but also the temporal information for each node.

Graph-level network embedding, unlike node embedding,  allows us to learn representations of entire graphs and directly compare different graphs, enabling investigation of fundamental graph ranking and retrieval problems such as the degree of similarity between graphs. Graph-level embedding methods have been studied extensively in the literature, but most of them focus on static networks \cite{narayanan2017graph2vec,chen2019gl2vec,tsitsulin2018netlsd,wang2021graph}. However, in real-world applications, dynamic networks are ubiquitous. To the best of our knowledge, only one prior method, called \texttt{tdGraphEmbed} \cite{tdgraph}, has been proposed for dynamic graph-level embedding. However, this method has a major limitation in that it treats dynamic graphs as a collection of independent static graph snapshots, ignoring the interactions between them.

To address this gap, we propose a novel method called the temporal backtracking random walk, which, when combined with the \emph{doc2vec} algorithm, can be used for dynamic graph-level embedding. Our method smoothly incorporates both graph structural and temporal information. We evaluate our method on five publicly available datasets for the task of temporal graph similarity ranking and demonstrate that it achieves state-of-the-art performance.

\section{Related Work}
In the introduction, we discussed \texttt{tdGraphEmbed} as the only existing method for dynamic graph-level embedding. In this section, we review two adjacent categories of graph embedding techniques: temporal node and static graph-level embedding.

Temporal node embedding methods differ from static node embedding methods such as \texttt{node2vec} \cite{node2vec}, \texttt{SDNE} \cite{SDNE}, and \texttt{GAE} \cite{GAE} in that they incorporate historical information to preserve both structural and temporal information. Matrix factorization techniques such as \texttt{TMNF} \cite{TFNM}, modified random walk algorithms such as \texttt{CTDNE} \cite{CTDNE}, and deep-learning-based methods such as \texttt{DynGEM} \cite{dyngem}, \texttt{dyngraph2vec} \cite{dyngraph2vec}, and variations like \texttt{DynAE}, \texttt{DynRNN}, and \texttt{DynAERNN} are examples of such techniques. Additionally, \texttt{DynamicTriad} \cite{DynamicTriad} employs the triadic closure process to develop closed triads from open triads.

For static graph-level embedding, various methods have been proposed, including the use of graph kernels (e.g., \texttt{graph2vec} \cite{narayanan2017graph2vec} employs graph kernels to extract features which are passed to a language model for embedding), random walks (\texttt{Sub2Vec} \cite{sub2vec}), multi-scale attention (\texttt{UGraphEmb} \cite{UGraphEmb}), and the Laplacian matrix and eigenvalues (e.g., \texttt{NetLSD} \cite{tsitsulin2018netlsd}).

\section{Approach}
In this section, we introduce our framework for the problem of representing each snapshot of a temporal graph as a low-dimensional vector that captures both the dynamic evolution information and graph topology.

\subsection{Problem Definition}
\label{sec:back}
Given a discrete temporal graph $G=(V, E, T)$, where each temporal edge $(u, v){t} \in E$ is directed from node $u$ to node $v$ at time $t \in T$, a snapshot of $G$ at time $t$ is defined as $G{t}=\left(V_{t}, E_{t}\right)$, which is the graph of all edges occurring at time $t$. The problem is to represent each snapshot $G_{t}$ as a low-dimensional vector $X_{t} \in \mathbb{R}^{n}$, where $n << |V|$, that captures both the dynamic evolution information and graph topology. We solve this problem in an unsupervised way and do not require any task-specific information.

\subsection{Our Framework}
Our framework consists of two parts: (1) building a multilayer graph and adopting temporal backtracking random walk on it (2) learning a \emph{doc2vec} language model on the output of the modified random walk to obtain graph-level embeddings.
First, we construct a multilayer weighted graph $M(V_{M},E_{M})$ that encodes the evolution between nodes. Each layer $M_{t}, t=0,1,\ldots,|T|$, is constructed by the nodes of $G$ and the edges of snapshot $G_{t}$. We build inter-layer edges between each pair of $M_{t}$ and $M_{t-1}$ by directly connecting the corresponding nodes from $t$ to $t-1$. Note that the edges between the two layers are unidirectional.
Next, we model each snapshot $G_{t}$ by using temporal backtracking random walk from each node as a sentence. Then all the sentences are concatenated to create a document representing the entire snapshot. During each step of the temporal backtracking walk, the walker can either stay in the current layer to obtain structural information or move to the previous layer to obtain historical evolving information. We define the \emph{stay} constant $\alpha$ such that the probability of staying in the current layer is $\alpha$, and the probability of going to the previous layer is $1-\alpha$. A temporal backtracking walk on $M$ is a sequence of vertices $\left\langle v_{1}, v_{2}, \cdots, v_{k}\right\rangle$ such that $\left\langle v_{i}, v_{i+1}\right\rangle \in E_{M}$ for $1 \leq i<k$, which can be derived by the transition probability on $M$.
Assuming that we have got $\left\langle v_{1}, v_{2}, \cdots, v_{i}\right\rangle$, and $v_{i} \in M_{t}$, the transition probability at step $i+1$ is defined as:
\begin{equation}
P\left(v_{i+1} |v_{i-1},v_{i}\right) = \begin{cases}

1-\alpha & v_{i+1} \in M_{t-1}\\
\frac{\alpha }{pZ} &  d_{v_{i-1}, v_{i+1}}=0 , v_{i+1} \in M_{t}\\ \frac{\alpha }{Z} &  d_{v_{i-1}, v_{i+1}}=1 , v_{i+1} \in M_{t} \\ \frac{\alpha }{qZ} &  d_{v_{i-1}, v_{i+1}}=2 , v_{i+1} \in M_{t}\\
0 & otherwise\end{cases}
\end{equation}
We draw inspiration from \texttt{node2vec} and introduce a modified version of the algorithm to capture temporal information. In this context, $d_{u,v}$ represents the length of the shortest path between node $u$ and $v$, while $p$ and $q$ are the return and in-out parameters, respectively. These parameters smoothly interpolate breadth-first and depth-first sampling. The normalizing constant $Z$ is also used. Alias sampling is used to perform each step of the temporal backtracking random walk in $O(1)$ time complexity.

The temporal backtracking random walk combines the proximity information of nodes within a layer with the structural information of previous timestamps. This approach is facilitated by the \emph{stay} constant, which is set to be larger than 0.5. This ensures that the influence of older timestamps decays smoothly as the probability of entering previous layers decreases exponentially.

We represent the context of each node in a $G_t$ snapshot as a sentence. These sentences are concatenated to create a document that represents the snapshot. As these sentences have no inherent order, we adopt a modified \emph{doc2vec} language model to learn a representation of the snapshot ``documents''. In this approach, each sentence is tagged with the corresponding timestamp ($t$ of $G_t$) as the paragraph id of \emph{doc2vec}. The final paragraph vector obtained after training is the dynamic graph-level embedding of $G_t$.

\begin{table*}[]
\small
\begin{tabular}{c|cccc|cccc}
\Xhline{2\arrayrulewidth} & \multicolumn{4}{c|} { Reddit - Game of Thrones } & \multicolumn{4}{c} { Reddit- Formula1 } \\
& p@10 & p@20 & $\tau$ & $\rho$ & p@10 & p@20 & $\tau$ & $\rho$ \\

\Xhline{2\arrayrulewidth} \textbf{Static graph-level embedding} & & & & & & & & \\
graph2vec & $0.260$ & $0.381$ & $0.038$ & $0.056$ & $0.169$ & $0.320$ & $0.043$ & $0.063$ \\
UGraphEmb & $0.278$ & $0.416$ & $0.046$ & $0.068$ & $0.238$ & $0.37$ & $0.026$ & $0.039$ \\
Sub2Vec & $0.160$ & $0.355$ & $0.022$ & $0.039$ & $0.182$ & $0.300$ & -$0.030$ & -$0.040$ \\
\Xhline{2\arrayrulewidth}\textbf{Temporal node-level embedding} & & & & & & & & \\
node2vec aligned & $0.336$ & $0.431$ & $0.069$ & $0.103$ & $0.214$ & $0.361$ & $0.047$ & $0.083$ \\
SDNE aligned & $0.352$ & $0.457$ & $0.120$ & $0.197$ & $0.262$ & $0.388$ & $0.044$ & $0.078$ \\
GAE aligned & $0.235$ & $0.342$ & $0.044$ & $0.066$ & $0.200$ & $0.342$ & $0.036$ & $0.062$ \\
DynGEM & $0.340$ & $0.441$ & $0.075$ & $0.113$ & $0.192$ & $0.339$ & $0.029$ & $0.045$ \\
DynamicTriad & $0.277$ & $0.364$ & $0.131$ & $0.195$ & $0.243$ & $0.396$ & $0.024$ & $0.033$ \\
DynAE & $0.192$ & $0.357$ & $0.019$ & $0.030$ & $0.229$ & $0.397$ & $0.009$ & $0.012$ \\
DynAERNN & $0.192$ & $0.349$ & -$0.002$ & -$0.004$ & $0.164$ & $0.357$ & $0.026$ & $0.037$ \\
\Xhline{2\arrayrulewidth}  \textbf{Temporal graph-level embedding} & & & & & & & & \\
 tdGraphEmbed & $0.355$ & $0.457$ & $0.160$ & $0.232$ & $\mathbf{0.274}$ & $0.400$ & $0.060$ & $0.092$ \\
  Our method & $\mathbf{0 .435}$ & $\mathbf{0 .481 }$ & $\mathbf{0 .177}$ & $\mathbf{0 .272}$ & $0 .265 $ & $\mathbf{0 .410 }$ & $\mathbf{0 .076 }$ & $\mathbf{0 .106 }$ \\
  \Xhline{2\arrayrulewidth} 
\end{tabular}
\centering
\caption{The temporal similarity results. The precision at K (p@K) metric is used to evaluate the method's accuracy. Additionally, we report Spearman's ($\rho$) and Kendall's ($\tau$) rank correlation coefficients to measure the method's consistency in ranking similar pairs of snapshots across different evaluation scenarios.}
\label{rank2}
\vspace{-10pt}
\end{table*}

\section{Experiments}

We evaluate the effectiveness of our dynamic graph-level embeddings by measuring their performance on the task of temporal graph similarity ranking. To this end, we use five publicly available datasets (Table \ref{dataset}) introduced by Beladev et al. \cite{tdgraph} and apply the same settings and metrics as used by them. Furthermore, we conduct scalability experiments to showcase our model's robustness and applicability to large networks commonly found in real-world applications.

 \begin{table}[htbp]
 \small
\begin{tabular}{ccc}
\Xhline{2\arrayrulewidth} Dataset & Nodes & Edges \\
\Xhline{2\arrayrulewidth} 
Reddit (Game of Thrones) & 156,732 & 834,753  \\
 Reddit (Formula1) & 38,702 & 254,731  \\
 Facebook wall posts & 46,873 & 857,815  \\
 Enron & 87,062 & $1,146,800$  \\
Slashdot & 51,083 & 140,778  \\
\Xhline{2\arrayrulewidth}
\end{tabular}
\centering
\caption{Dataset statistics (see \cite{tdgraph} for more detail).}  
\label{dataset}
\vspace{-25pt}
\end{table}

\begin{table*}[]
\small
\begin{tabular}{c|p{6mm}p{6mm}cc|p{6mm}p{6mm}cc|p{6mm}p{6mm}cc}
\Xhline{2\arrayrulewidth} &\multicolumn{4}{c|} { Enron }  & \multicolumn{4}{c|} { Facebook-wall posts } & \multicolumn{4}{c} { Slashdot } \\
& p@10 & p@20 & $\tau$ & $\rho$ & p@10 & p@20 & $\tau$ & $\rho$ & p@5 & p@10 & $\tau$ & $\rho$ \\

\Xhline{2\arrayrulewidth} \textbf{Static graph-level embedding} & & & & & & & & \\
graph2vec & $.045$ & $.059$ & -$.033$ & -$.046$ & $.423$ & $.713$ & $.120$ & $.176$ & $.292$ & $.800$ & $.026$ & $.045$ \\
UGraphEmb & $.168$ & $.269$ & $.110$ & $.150$ & $. 7 5 0$ & $.871$ & $.355$ & $.452$ & $.462$ & $.900$ & $.215$ & $.271$ \\
Sub2Vec & $.073$ & $.137$ & $.028$ & $.044$ & $.353$ & $.685$ & $.012$ & $.021$ & $.385$ & $.808$ & $.037$ & $.074$ \\
\Xhline{2\arrayrulewidth}\textbf{Temporal node-level embedding} & & & & & & & & \\
node2vec aligned & $.379$ & $.452$ & $.107$ & $.139$ & $.680$ & $.840$ & $.303$ & $.414$ & $.538$ & $.908$ & $.229$ & $.306$ \\
SDNE aligned & $.316$ & $.400$ & $.087$ & $.138$ & $.400$ & $.645$ & $.095$ & $.120$ & $.415$ & $.885$ & $.095$ & $.124$ \\
GAE aligned & $.277$ & $.360$ & $.118$ & $.156$ & $.613$ & $.820$ & $.292$ & $.397$ & $.492$ & $.885$ & $.168$ & $.227$ \\
DynGEM & $.335$ & $.377$ & $.103$ & $.143$ & $.356$ & $.733$ & $.094$ & $.115$ & $.569$ & $\mathbf{ . 9 1 5}$ & $.245$ & $.314$ \\
DynamicTriad & $.322$ & $.425$ & $.112$ & $.153$ & $.733$ & $.818$ & $.271$ & $.395$ & $.646$ & $.869$ & $.201$ & $.276$ \\
DynAE & $.069$ & $.145$ & $.009$ & $.012$ & $.389$ & $.743$ & $.122$ & $.163$ & $.473$ & $.900$ & $.002$ & $.025$ \\
DynAERNN & $.061$ & $.110$ & $.004$ & $.006$ & $.393$ & $.755$ & $.065$ & $.076$ & $.509$ & $.900$ & $.041$ & $.088$ \\
\Xhline{2\arrayrulewidth}  \textbf{Temporal graph-level embedding} & & & & & & & & \\
 tdGraphEmbed & $. 3 8 5$ & $ . 4 8 9$ & $ . 1 2 7$ & $ . 1 8 8$ & $. 7 5 0$ & $  . 8 9 2$ & $. 3 9 8$ & $ . 5 2 2$ & $\mathbf{ . 7 8 5}$ & $\mathbf{ . 9 1 5}$ & $. 3 4 7$ & $ . 4 6 3$ \\
   Our method & $\mathbf{ .479}$ & $\mathbf{ .532 }$ & $\mathbf{ .172}$ & $\mathbf{ .251}$ & $\mathbf{ .806 }$ & $\mathbf{.896} $ & $\mathbf{ .447 }$ & $\mathbf{ . 559}$ & $ .723 $ & $ .885 $ & $\mathbf{ .400 }$ & $\mathbf{ .524 }$ \\
\Xhline{2\arrayrulewidth}

\end{tabular}
\centering
\caption{Continuation of Table \ref{rank2}}
\label{rank3}
\vspace{-15pt}
\end{table*}

\vspace{-10pt}
\subsection{Experimental Setup}
We compare our model with three types of baselines: static graph-level embedding methods (represented by \texttt{graph2vec}, \texttt{UGraphEmb}, and \texttt{Sub2vec}), temporal node-level embedding methods (represented by \texttt{node2vec aligned}, \texttt{SDNE aligned}, \texttt{GAE aligned}\footnote{Here, the term ``aligned'' means that each snapshot is trained separately, and the embeddings are then rotated for alignment \cite{hamilton2016diachronic}. Since these three methods are static, we use them to represent temporal node-level embeddings.}), and temporal graph-level embedding methods (represented by \texttt{DynGEM}, \texttt{DynamicTriad}, \texttt{DynAE}, \texttt{DynAERNN}, and the only existing state-of-the-art method, \texttt{tdGraphEmbed}). For all baselines, we use the same parameter settings as introduced by Beladev et al. and report the best results between our experiments and the results reported by them. This is done to ensure fairness and to err on the side of caution. 

For our model, we set the number of temporal backtracking random walks from each node to 40, with a length of 32. We set the return parameter $p$ to 1, the in-out parameter $q$ to 0.5, and the \emph{stay} constant $\alpha$ to 0.8. For the \emph{doc2vec} model training, we set the maximum distance between the current and predicted word within a sentence to 5, the initial learning rate to 0.025, and the size of the final embedding to 128.

\subsection{Temporal Similarity Ranking}
This task aims to test a model's ability to capture the similarity among each snapshot of a dynamic graph $G$. For a given snapshot $G_t$, the most similar snapshot to it may not be its immediate neighbors $G_{t-1}$ or $G_{t+1}$, but some other snapshot that is far away from it \cite{tdgraph}. The temporal similarity ranking task has numerous potential real-world applications. For example, it can be used to detect organized influence operations on social media by analyzing the similarity of dynamic share/reply networks.

To evaluate our model, we train it to obtain representations for all the snapshots in five publicly available datasets introduced by Beladev et al. \cite{tdgraph}, using the same settings and metrics as their work. We compare our model with three types of baselines: static graph-level embedding (represented by \texttt{graph2vec}, \texttt{UGraphEmb}, and \texttt{Sub2vec}), temporal node-level embedding (represented by \texttt{node2vec aligned}, \texttt{SDNE aligned}, \texttt{GAE aligned}, \texttt{DynGEM}, \texttt{DynamicTriad}, \texttt{DynAE}, and \texttt{DynAERNN}), and the only existing state-of-the-art method for temporal graph-level embedding, \texttt{tdGraphEmbed}. For each snapshot $G_t$, we rank all the other snapshots $G_i, (i\neq t)$ based on the cosine similarity between their embeddings $X_{t}$ and $X_{i}$: $cos({X}{t},{X}{i})=\frac{{X}{t} \cdot {X}{i}}{|{X}{t}||{X}{i}|}$. We then use the predicted and ground truth ranking lists of $G_t$ to calculate the average precision at 10 and 20, and Spearman's and Kendall's rank correlation coefficients ($\rho$ and $\tau)$. For the Slashdot dataset, we report precision at 5 and 10 since there are only 13 time-steps.

Our model outperforms all the baselines for all the experiments, except for three cases (out of 220) where \texttt{tdGraphEmbed} performs best, as shown in Tables \ref{rank2} and \ref{rank3}. We also conduct scalability experiments to demonstrate our model's robustness and applicability to large networks commonly found in real-world applications.

\begin{figure}[]
\vspace{-25pt}
\centerline{\includegraphics[width=0.25\textwidth]{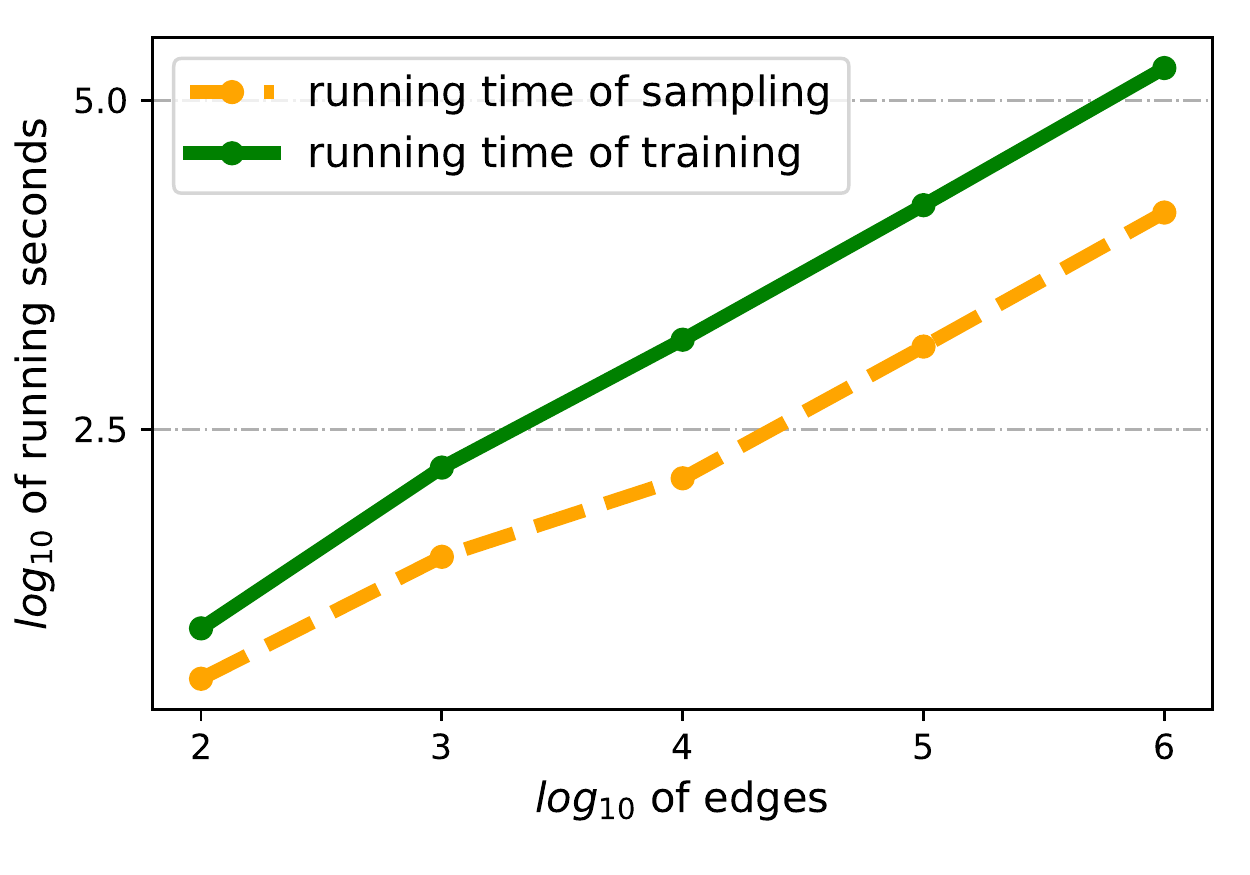}}
\caption{Scalability experiment results on Erdos-Renyi graphs with an average degree of 10.}
\label{fig:sca}
\vspace{-20pt}

\end{figure}

\subsection{Scalability Analysis}
To evaluate the scalability of our proposed model, we conduct experiments on Erdos-Renyi graphs with increasing sizes from 100 to 1,000,000 edges, where each node has an average degree of 10. We uniformly split the edges of each graph into 10 different snapshots and learn the temporal graph representations using our model with default parameters. The experiments are conducted on a Lambda Deep Learning 2-GPU Workstation (RTX 2080). As shown in Figure \ref{fig:sca}, the log-log plot of the running time versus the number of nodes demonstrates that our model's performance is polynomial in time with respect to the graph's size. The slopes of the curves are less than 1 in the log-log space, indicating that our method performs in sub-linear time due to its use of parallel processing. Thus, our proposed method can be efficiently scaled to handle large networks commonly found in real-world applications.

\section{Conclusion}
\vspace{-2pt}
We introduced a novel dynamic graph-level embedding method based on temporal backtracking random walk. Our method can capture both the structural and evolving information of dynamic graphs. Experimental results on five publicly available datasets for temporal graph similarity ranking show the superiority of our proposed method over several baselines. Moreover, our model is scalable to larger networks, which makes it applicable to real-world scenarios. Our method provides a promising solution for dynamic graph embedding tasks and can be applied to various real-world applications.
\vspace{-15pt}

\bibliographystyle{ACM-Reference-Format}
\bibliography{sample-base}
\end{sloppypar}
\end{document}